# Causality-Based Feature Importance Quantifying Methods—PN-FI, PS-FI and PNS-FI

Shuxian Du
a1240320952@163.com
Harbin Engineering University

Yaxiu Sun
sunyaxiu@hrbeu.edu.cn
Harbin Engineering University

Changyi Du
0833scls@163.com
Chengdu No. 7 High School

## *Abstract*

*In the current ML field models are getting larger and more complex, and data used for model training are also getting larger in quantity and higher in dimensions. Therefore, in order to train better models, and save training time and computational resources, a good Feature Selection (FS) method in the preprocessing stage is necessary. Feature importance (FI) is of great importance since it is the basis of feature selection. Therefore, this paper creatively introduces the calculation of PN (the probability of Necessity)，PN (the probability of Sufficiency), and PNS (the probability of Necessity and Sufficiency) of Causality into quantifying feature importance and creates 3 new FI measuring methods—— PN-FI, which means how much importance a feature has in image recognition tasks, PS-FI that means how much importance a feature has in image generating tasks, and PNS-FI which measures both. The main body of this paper is three RCTs, with whose results we show how PS-FI, PN-FI, and PNS-FI of 3 features——dog nose, dog eyes, and dog mouth are calculated. The experiments show that firstly, FI values are intervals with tight upper and lower bounds. Secondly, the feature dog eyes has the most importance while the other two have almost the same. Thirdly, the bounds of PNS and PN are tighter than that of PS's.*

## 1. INTRODUCTION

Feature importance (FI) is of great importance since it is the basis of Feature Selection (FS), which has become increasingly important for data analysis, machine learning, and data mining [1].

Nowadays, data from the real world are often large in quantity and high in dimensions. Therefore, a good FI measurement method will give rise to a good FS method, which contributes a lot. The first is that finding the useful features and casting away irrelevant and redundant features in the preprocessing stage will save us a lot of training time and computational resources, and the second is that it can prevent the model from overfitting [2]. Besides, we can get a deeper understanding of the model by looking at the feature set it chooses.

In current papers, FS methods are divided into 3 categories——filter-based, wrapper-based, and embedded-based FS.

Wrapper-based methods evaluate attribute sets using a machine learning method via an iterative search process [3]. Ensemble learning is based on combining multiple same or different models instead of a single model to solve a particular problem, the most commonly used is Random Forest (RF) [4].

Filter-based feature selection is to select a subset of relevant features from a larger set of features, in which "filter-based" means that this method involves filtering and ranking features based on some criteria. Here are some commonly used filters:

- Filter variance performs for each feature an analysis of variance [5].

- Filter Kruskal–Wallis applies for each feature a Kruskal–Wallis rank sum test which is the non-parametric equivalent of the analysis of variance [6].
- Filter chi-squared measures the independence between a categorical feature and a categorical target variable [7].
- Filter mutual information measures the amount of information that a feature provides to the target variable [8].

**In this paper, we creatively use PN, PS, and PNS in Causality to present a simple way to quantify the Feature Importance (FI) in image recognition (IR), giving rise to PNS-FI, that is, PNS-feature-importance. Moreover, in the near future we will present a new filter——PNS filter.**

The main body of this paper is 3 RCTs (Random Controlled Trial).

CNN is used to calculate the FI importance of 3 chosen features because its dataset-splitting strategy makes RCTs easier.

To accomplish these experiments, 2 AMD EPYC 7R32 48-Core Processor 2.80 GHz CPUs and a NVIDIA GeForce RTX 3080 Ti are used.

## 2. PRELIMINARIES:

This paper is written under the structure of Judea Pearl's **SCM (Structural Causal Model)**. SCM is a framework for representing and analyzing causal relationships among variables in a system [9-25].

Under this structure, firstly, x' means that the thing x does not happen. Secondly, the counterfactual sentence, "Variable Y would have the value y, had X been x" is denoted as $Y_x = y$, and shorted as $y_x$. Thirdly, the probabilities derived from observation data are expressed as joint probability functions like P(x|y). On the other hand, the probabilities derived from experimental data (such as RCT) are expressed as causal probabilities $P(y_x)$ [26].

Then, several definitions and theorems are needed [27,28].

**Definition 1 (Probability of necessity (PN))**

Let X and Y be 2 binary variables in a causal model M, let x and y stand for the propositions X=true and Y=true, respectively, and x' and y' for their complements. The probability of necessity is defined as the expression:

$$PN = P(Y_{x'} = false \mid X = true, Y = true)$$
$$= P(y'_{x'} \mid x, y) \quad (1)$$

In other words, what is the probability that x is the reason for y. If PS=1, then x is 100% a reason for y.

**Definition 2** (Probability of sufficiency (PS))

$$PS = P(y_x \mid x', y') \quad (2)$$

In other words, to which extent does x decide the happening of y. If PS=1, then x is the only reason for y.

**Definition 3** (Probability of necessity and sufficiency (PNS))

$$PNS = P(y_x, y'_{x'}) \quad (3)$$

The PNS encodes the extent to which a certain treatment to a particular outcome would be both necessary and sufficient, that is, the probability that Y would respond to X in both of the ways described above [13].

PN, PS, and PNS are together called **the Probabilities of Causation**.

**Definition 4** (Monotonicity)

A variable Y is said to be monotonic relatively to variable X in a causal model M iff:

$$y'_x \wedge y_{x'} = false \qquad (4)$$

**Definition 5** (Exogeneity)

A Variable is said to be exogenous for Y in model M iff:
$$P(y_x) = P(y/x) \quad and \quad P(y_{x'}) = P(y/x') \qquad (5)$$

This definition is also called the "no confounding" condition, which means that **without confounding, the causal probabilities can be reduced to conditional probabilities.**

**Theorem 1** Under the condition of exogeneity, the three probabilities of causation are bounded as follows:

$$\begin{aligned} \max[0, P(y/x) - P(y/x')] &\leq PNS \leq \min[P(y/x), P(y'/x')] \\ \frac{\max[0, P(y/x) - P(y/x')]}{P(y/x)} &\leq PN \leq \frac{\min[P(y/x), P(y'/x')]}{P(y/x)} \\ \frac{\max[0, P(y/x) - P(y/x')]}{P(y'/x')} &\leq PN \leq \frac{\min[P(y/x), P(y'/x')]}{P(y'/x')} \end{aligned} \qquad (6)$$

## 3. THE CALCULATION STRATEGIES OF PN-FI, PS-FI AND PNS-FI

In this part, PN-FI, PS-FI, and PNS-FI of 3 features are calculated.

**PN-FI of a feature means how much importance the feature has in image recognition tasks, PS-FI means how much importance the feature has in image generating tasks, and PNS-FI measures both.**

The chosen features are dog nose, dog eyes, and dog mouth, and the task is to recognize dogs from 11 kinds of animals.

Two variables are involved——**x means the feature can be seen, and x' means the feature cannot be seen. y means a dog sample is recognized while y' means the dog sample is not recognized.**

The dataset is handmade, containing **6000 raw animal pictures**, which is composed of part of **Stanford Dogs, Kaggle Animal-10, and some treated pictures**. It includes **3000 dogs** and **3000 other 10 kinds of animals**. Among these 3000 dogs, only 31 of them have their noses covered, 23 have their eyes covered and 45 have their mouths covered. The 10 other kinds of animals are cats, spiders, horses, cattle, goats, butterflies, chickens, lions, elephants, and squirrels.

Since nose/eyes/mouth-covered samples are too few, in order to make the effect more obvious, we have some dog samples' noses, eyes, or mouths manually covered. Then we have 600 "no-eyes", 600 "no-nose" and 600 "no-mouth" dog samples.

To wipe out the confounding of the color, shape, and texture of the covering, the "no-eyes", "no-nose" and "no-mouth" dog samples are processed like in Figure 1 below:

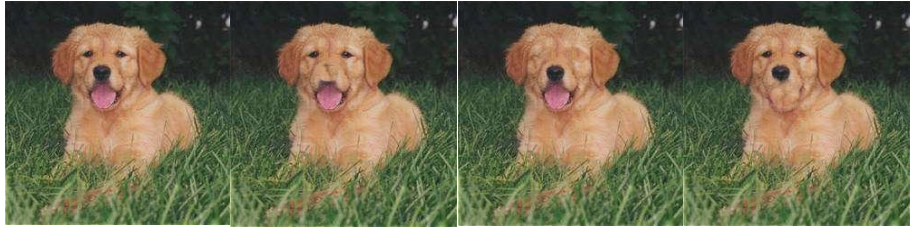

Figure 1. The original dog sample and the "no-nose", "no-eyes", "no-mouth" dog samples.

**Therefore, Theorem 1(Exogeneity) in preliminaries is applicable in this situation.**

### 3.1 RCTs to Gain Causal Probabilities

In this part, 3 RCTs, one for each feature, are carried out to obtain causal probabilities and then to calculate the probabilities of causation.

What is important is that the data in training and validation sets are randomly assigned whilst the test set is controlled, which means in the controlled group it is the raw pictures while in the experimental group, it is the same batch of pictures as in the controlled group but treated as in Figure 1.

Therefore, the composition of the dataset used for three experimental trials is shown in TABLE 1, and the one used for three controlled trials is shown in TABLE 2. For convenience, feature+ means the feature is revealed, whilst feature- stands for being covered. The dataset split strategy is the classic 6:2:2.

TABLE1. Data assignment for experimental groups, and **the order is nose/eyes/mouth**.

|  | feature+ | feature- | non-dog |
|---|---|---|---|
| training set | 1774/1781/1766 | 26/19/34 | 1800/1800/1800 |
| validation set | 595/596/589 | 5/4/11 | 600/600/600 |
| test set | 0/0/0 | 600/600/600 | 600/600/600 |

TABLE 2. Data assignment for controlled groups, and **the order is nose/eyes/mouth**.

|  | feature+ | feature- | non-dog |
|---|---|---|---|
| training set | 1774/1781/1766 | 26/19/34 | 1800/1800/1800 |
| validation set | 595/596/589 | 5/4/11 | 600/600/600 |
| test set | 600/600/600 | 0/0/0 | 600/600/600 |

The CNN we use for recognition is 10 layers in depth, including 8 deep layers and 2 fully connected layers, the optimization algorithm is Adam, the batch size for 3 sets is 20 and the training epochs is 100. The training results are shown in TABLE 3 below.

TABLE 3. The RCT results——the **true-positive rates** and **false-positive rates**.

| feature | group | true-positive rate | false-positive rate |
|---|---|---|---|
| dog nose | experimental | 75% | 25% |
|  | controlled | 84% | 16% |
| dog eyes | experimental | 62.5% | 37.5% |
|  | controlled | 81% | 19% |
| dog mouth | group | 69% | 31% |
|  | controlled | 77.5% | 22.5% |

With the data in TABLE 3, the causal probabilities can be obtained, which are presented in TABLE 4.

TABLE 4. The **causal probabilities** gained from RCTs.

|  | $P(y_x)$ | $P(y_{x'})$ | $P(y'_x)$ | $P(y'_{x'})$ |
|---|---|---|---|---|
| dog nose | 0.84 | 0.75 | 0.16 | 0.25 |
| dog eyes | 0.81 | 0.625 | 0.19 | 0.375 |
| dog mouth | 0.775 | 0.69 | 0.225 | 0.31 |

## 3.2 Calculating PN-FI, PS-FI and PNS-FI

From TABLE 4 we can easily see that **Definition 4** is not met, and because of our delicate preprocessing of the pictures, the no-confounding condition, that is, exogeneity of **Definition 5** is met, then the causal probabilities can be written as conditional probabilities as below.

$$P(y_x) = P(y/x)$$
$$P(y_{x'}) = P(y/x')$$
$$P(y'_x) = P(y'/x)$$
$$P(y'_{x'}) = P(y'/x')$$
(7)

Then **Theorem 1** is applicable, the PN-FI, PS-FI, and PNS-FI can be bounded as follows.

$$\max[0, P(y/x) - P(y/x')] \leq PNS \leq \min[P(y/x), P(y'/x')]$$
$$\frac{\max[0, P(y/x) - P(y/x')]}{P(y/x)} \leq PN \leq \frac{\min[P(y/x), P(y'/x')]}{P(y/x)}$$
$$\frac{\max[0, P(y/x) - P(y/x')]}{P(y'/x')} \leq PN \leq \frac{\min[P(y/x), P(y'/x')]}{P(y'/x')}$$
(8)

The results are shown in TABLE 5 below.

TABLE 5. The PN-FI, PS-FI and PNS-FI of 3 features.

|  | PN-FI | PS-FI | PNS-FI |
| --- | --- | --- | --- |
| dog nose | [0.107, 0.30] | [0.36, 1] | [0.09, 0.25] |
| dog eyes | [0.228, 0.463] | [0.493, 1] | [0.185, 0.375] |
| dog mouth | [0.110, 0.4] | [0.274, 1] | [0.085, 0.31] |

PN-FI of dog nose/eyes/mouth means how much importance dog nose/eyes/mouth has in **recognizing** a dog sample, and PS-FI means how much importance dog nose/eyes/mouth has in **generating** a dog sample.

## 4. CONCLUSION AND PROSPECT

This paper creatively introduces the calculation of **PN (the probability of Necessity), PN (the probability of Sufficiency), and PNS (the probability of Necessity and Sufficiency)** of **Causality** into quantifying feature importance and creates three new feature importance measuring methods—— **PN-FI**, which means how much importance a feature has in image recognition tasks, and **PS-FI** which means how much importance a feature has in image generating tasks, and **PNS-FI** which measures both. Besides, Section 3 shows how to obtain causal probabilities through three RCTs and how to calculate PN-FI, PS-FI, and PNS-FI with the causal probabilities.

The experiments show that firstly, FI values are intervals with tight upper and lower bounds. Secondly, the feature dog eyes has the most importance while the other two have almost the same. Thirdly, the bounds of PNS and PN are tighter than that of PS's.

In the near future we will continue our research in three steps:

Firstly, **we will create a new filter called PNS-Filter based on PN-FI, PS-FI and PNS-FI we proposed.**

Secondly, **we will find a proper empirical threshold between relevant and irrelevant features, redundant features**, and **compare our filter with other existing filters like variance and mutual information on classification accuracy and running time**.

Thirdly, we will try to **gain PN-FI, PS-FI, and PNS-FI values with tighter bounds**.

References:

[1]. Andrea Bommerta, Xudong Sun, Bernd Bischl, Jörg Rahnenführer, Michel Lang, "Benchmark for filter methods for feature selection in high-dimensional classification data", Computational Statistics & Data Analysis, Volume 143, March 2020.

[2]. Saúl Solorio-Fernández, J. Ariel Carrasco-Ochoa, José Fco, Martínez-Trinidad, "A review of unsupervised feature selection methods", ArtificialIntelligenceReview(2020)53:907–948.

[3]. Haidi Rao, Xianzhang Shi, Ahoussou Kouassi Rodrigue, etc., "Feature selection based on artificial bee colony and gradient boosting decision tree", Applied Soft Computing Journal 74 (2019) 634–642.

[4]. Vasilii Feofanov, Emilie Devijver, Massih-Reza Amini, "Wrapper feature selection with partially labeled data", Applied Intelligence volume 52, pages12316–12329 (2022)

[5]. Xiaofei He; Ming Ji; Chiyuan Zhang; Hujun Bao, "A Variance Minimization Criterion to Feature Selection Using Laplacian Regularization", IEEE Transactions on Pattern Analysis and Machine Intelligence ( Volume: 33, Issue: 10, October 2011)

[6]. Kruskal, W.H., Wallis, W.A., "Use of ranks in one-criterion variance analysis", J. Amer. Statist. Assoc. 1952.47 (260), 583–621.

[7]. Rasch, D., Kubinger, K.D., Yanagida, T., "Statistics in Psychology using R and SPSS", 2011, John Wiley & Sons, Inc., Hoboken, NJ, USA.

[8]. Brown, G., Pocock, A., Zhao, M.-J., Luján, M., "Conditional likelihood maximization: A unifying framework for information theoretic feature selection". 2012, J. Mach. Learn. Res. 13, 27–66.

[9]. Pearl J. "The Book of Why", 2019.

[10]. Judea Pearl and Ang Li, "Unit selection with causal diagram", arxiv2022.

[11]. Pearl J. "Theoretical impediments to machine learning with seven sparks from the causal revolution", arXiv, 2018.

[12]. Pearl J. "Causal Inference In Statistics:A Primer", Wiley, 2016.

[13]. Bernhard Schölkop, "Challenging Common Assumptions in the Unsupervised Learning of Disentangled Representations", ICML 2019.

[14]. S Mueller, J Pearl, "Personalized decision making–A conceptual introduction", Journal of Causal Inference, 2023.

[15]. Bernhard Schölkopf, "Causality for Machine Learning",ACM, 2022.4.

[16]. Pearl J. "The seven tools of causal inference, with reflections on machine learning", Communications of theACM, Vol. 62, No. 3, 2019

[17]. Ang Li and Pearl.J, "Probabilities of Causation with Nonbinary Treatment and Effect", arXiv preprint arXiv:2208.09568, 2022

[18]. E Bareinboim, JD Correa, D Ibeling, T Icard, "On Pearl's hierarchy and the foundations of causal inference Probabilistic and causal inference: the works of judea pearl, 2022.

[19]. Ang Li and Pearl.J, "Unit Selection Based on Counterfactual Logic", Revised version to appear in the Proceedings of IJCAI 2019.

[20]. B. Schölkopf et al., "Toward Causal Representation Learning," in Proceedings of the IEEE, vol. 109, no. 5, pp. 612-634, May 2021, doi: 10.1109/JPROC.2021.3058954.

[21]. Siyuan Guo, Viktor Tóth, Bernhard Schölkopf, Ferenc Huszár. "Causal de Finetti: On the Identification of Invariant Causal Structure in Exchangeable Data", arxiv2203.15756.

[22]. Ang Li and Pearl.J, "Unit Selection with Causal Diagram", arXiv:2109.07556v1 [cs.AI] 15 Sep 2021

[23]. Giambattista Parascandolo, Niki Kilbertus, Mateo Rojas-Carulla, Bernhard Schölkopf, "Learning Independent Causal Mechanisms", In Proceedings of the 35th International Conference on Machine Learning, PMLR

80:4036-4044, 2018.

[24]. Mengyue Yang, Furui Liu, "CausalVAE: Disentangled Representation Learning via Neural Structural Causal Models", CVPR2021 open access.

[25]. A Li, J Pearl, "Bounds on causal effects and application to high dimensional data", In Proceedings of the AAAI Conference on Artificial Intelligence, 2022.

[26]. Pearl J. "Causality: Models, Reasoning, and Inference", 2000.

[27]. Jin Tian and Judea Pearl, "Probabilities of Causation: Bounds and Identification", UNCERTAINTY IN ARTIFICIAL INTELLIGENCE PROCEEDINGS 2000.

[28]. A Li, S Mueller, J Pearl, "Epsilon-Identifiability of Causal Quantities" arXiv preprint arXiv:2301.12022, 2023.8.27.